\documentclass[conference]{IEEEtran}
\usepackage{times}
\usepackage{amsmath}
\usepackage{graphicx}
\usepackage{helvet}
\usepackage{sansmath}

\usepackage[numbers]{natbib}
\usepackage{multicol}
\usepackage[bookmarks=true]{hyperref}

\pdftrailerid{}
\hypersetup{pdfauthor={}, pdftitle={}, pdfsubject={}, pdfkeywords={}, pdfcreator={}, pdfproducer={}}

\begin{document}

\title{Venture: Long-Term Exploration with a Novelty Reward for Vision-Language-Action Policies
\unskip}

\author{\authorblockN{Mehmet Turan Yard{\i}mc{\i}}
\authorblockA{Karlsruhe Institute of Technology}}

\maketitle

\begin{abstract}
A robot may work in the same environment for years. Over that time it will meet tasks that its pretraining never covered. Online reinforcement learning (RL) can teach such tasks from the robot's own experience. The usual way to explore injects noise into the process that samples the actions. This noise only varies what the robot's policy already does. It refines known skills and rarely finds a solution that needs many new steps. We add a second kind of exploration: a small novelty term in the reward. The policy is a vision-language-action (VLA) model. It computes three signals about itself. They are the novelty of what it sees, its competence and the entropy of its sampled actions. It reads its competence from its own estimate of future reward. These signals set the scale of the injected noise, and the novelty signal also forms the novelty term in the reward. We call the method Venture. The noise explores in the short term. The novelty reward explores in the long term, toward states the policy has not visited. We study this with GR00T N1.7, a pretrained VLA policy, on simulated manipulation tasks. The tasks come in four groups of ten, plus a pool of 90 further tasks. Four released policies, each fine-tuned on one group, solve 74 to 95 of 100 episodes of their own group. On each other group they solve 0 to 2 of 100. Tasks outside a policy's own group thus mostly start at zero success, and we test exploration on such tasks. For one of these policies, we fitted the network behind the novelty signal on episodes of the tasks it solves. With probability 0.82 to 0.84, this network scores an episode of a task with zero success as more novel than an episode of a solved task. For a second VLA policy, $\pi_{0.5}$, the probability is 0.95 to 0.97 within the pool of 90 further tasks. Chance is 0.5. Our main hypothesis is that the novelty reward raises success on tasks with zero success at the start. A second hypothesis is that the solved tasks stay solved. A first pair of RL runs, with one training seed each, shows no gain on the tasks with zero success. Both runs start from the policy fine-tuned on one group and train on a pool of 90 further tasks. The ten tasks of that group stay out of training. On these ten tasks the starting policy solves 74 of 100 episodes. At the end of training, the run with the reward solves 75 of them and the run without it 66. Between the two runs, this difference is within chance (p = 0.12). The main evaluation of the full paper uses two household benchmark suites, LocoManiLab and RobotAgentBench; LIBERO serves only for fast preliminary tests.

\end{abstract}

\IEEEpeerreviewmaketitle

\section{Introduction}\label{sec:1}

\begin{figure*}[t]
\centering
\includegraphics{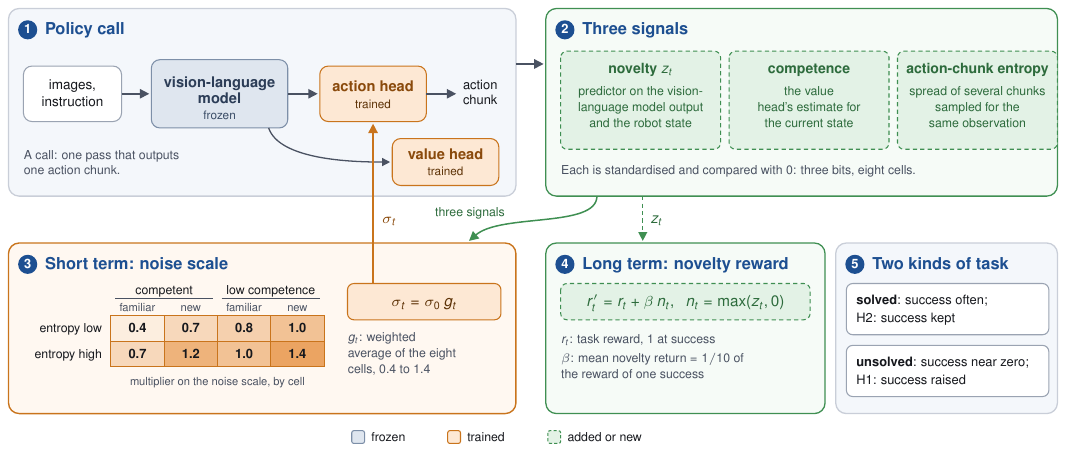}\par\vspace{\baselineskip}
\caption{Overview of Venture. (1) One call of the policy outputs one action chunk. (2) Three signals are computed from the policy's own activations and outputs. (3) Together they set the noise scale of the action head. (4) The novelty signal is also added to the task reward. (5) The hypotheses concern solved and unsolved tasks.}
\label{fig:method}
\end{figure*}

A household robot gets a new appliance, an air fryer. The data its model was trained on did not contain this device. The next model update is six months away. The robot has to learn to use the device before then, from its own experience.

Fine-tuning by online reinforcement learning (RL) can do this. A pretrained vision-language-action (VLA) policy acts, receives a reward only when it succeeds, and improves. Many VLA policies produce actions by flow matching: they move random noise along a learned velocity field until it becomes an action. Such a policy is made stochastic by noise injected into this sampling process. It is then trained with a policy-gradient method~\cite{2510.25889,2505.22094}. Such a method raises the probability of actions that led to reward. The injected noise is the exploration.

Noise injection is short-term exploration. Each sample moves the action a little around what the policy would have done. This adapts existing skills. A task the policy has never solved often needs a new sequence of many steps. Small random changes around the old behaviour rarely produce such a sequence. On such a task the reward stays at zero, and learning starts only with the first success. In one study of a flow-matching VLA, RL with noise alone rescued 10 of 27 tasks that started at zero success. It left 17 at zero and drove 3 partly solved tasks to zero~\cite{2609.21650}.

People explore differently: they try new ways of acting whose payoff they cannot know in advance. A child has only seen towers built one block at a time. Past some height, a single block on one side tips the tower over. The child tries placing two blocks at once, one on each side, and the tower stays balanced and grows. Nobody demonstrated this to the child, and small variations of the demonstrated way do not reach it.

We therefore add long-term exploration to the reward. A novelty term pays the policy for reaching states it has not seen. The policy then learns to go there on purpose. It can also combine what it learned on other tasks into new attempts. Noise injection stays, so the policy explores in both ways at once. Undirected noise and a directed reward have been combined in general RL~\cite{2412.12098}. We carry the combination into on-policy RL of a VLA policy. On-policy RL learns from rollouts, the episodes that the current policy has just played.

A task can be new to the policy in two ways. In the first, the policy has seen the task rarely or not at all. In the second, it has seen something similar but has found no way to solve the task.

The need for long-term exploration grows as the robot's body gains arms, legs and joints. A second arm adds options: one hand can hold an object while the other works on it. Legs, a bending torso and a shifting centre of mass add further options.

Contributions:

\begin{enumerate}
\item Two time scales of exploration in on-policy RL of a VLA policy: injected noise for the short term, a novelty reward for the long term.
\item A method built on three signals that the policy computes about itself: novelty, competence and action entropy. All three set the noise scale each time the policy computes its next actions, and novelty also enters the reward.
\item An evaluation design on two household benchmark suites, LocoManiLab and RobotAgentBench, three hypotheses, and preliminary measurements of the three signals and of the policy's success before RL.
\end{enumerate}

\section{Related work}\label{sec:2}

RL fine-tuning of flow-matching policies. ReinFlow~\cite{2505.22094} makes a flow-matching policy stochastic with a learned noise network. $\pi_{\mathrm{RL}}$~\cite{2510.25889} offers two ways: a learned noise network, or a sampling equation changed from deterministic to stochastic. In both works, exploration comes from the injected noise, and the reward has no exploration term.

Signals of the policy's own uncertainty. FIPER~\cite{2510.09459} predicts at run time that a policy is about to fail, from two scores. One is random network distillation on the policy's embeddings: a predictor network learns to match a fixed random network, and its error marks unfamiliar inputs~\cite{1810.12894}. The other is the entropy across action chunks, the short sequences of actions that the policy outputs at once. FIPER trains its policies by imitation and uses the scores only at run time. In a comparison of uncertainty measures for flow-matching policies, random network distillation flags new situations even when the policy succeeds in them~\cite{2606.18043}.

A controller for exploration noise~\cite{2609.28838} sets the noise scale per step from two signals, novelty and competence. It reads competence from value networks, which estimate the expected return. It combines two readings: competence counts as low when the estimate of a single network is low or when the disagreement within an ensemble is high. It leaves the reward unchanged. Like that controller, we read novelty and competence. We add the action-chunk entropy as a third signal and also use novelty in the reward. In another work the policy's confidence selects reference trajectories, and similarity to them serves as a reward at test time~\cite{2606.29892}.

Rewards the agent computes for itself. Random network distillation~\cite{1810.12894} and curiosity~\cite{1705.05363} pay the agent for its prediction error. In TacEx~\cite{2609.40134}, a separate agent steers the input noise of a frozen VLA policy. The agent receives a bonus for state-action pairs on which an ensemble of models disagrees, with the tactile part of the disagreement weighted. EvoVLA~\cite{2511.16166} adds a curiosity reward to on-policy fine-tuning of a VLA policy. It computes the curiosity from the relative pose of gripper and object. We compute our bonus from the robot state and from the output of the policy's own vision-language model, the part that reads the camera images and the instruction. DRIVE~\cite{2610.09943} adds a diversity bonus to on-policy RL of a flow-matching VLA, and the bonus is zero for every rollout that fails. Our bonus is paid before any success. Prompt Driven Exploration~\cite{2607.08837} explores by rewriting the instruction. As a baseline it reports a novelty bonus. On the simulated manipulation benchmark LIBERO-PRO, the bonus reaches 55.5 percent success over all tasks, against 58.4 percent for action noise. On the tasks that the work calls hard, the bonus reaches 10.5 percent and action noise 13.3 percent. In our setting, conditions (a) and (e) of Section~\ref{sec:5} make the corresponding comparison: fixed noise with and without our novelty reward.

Exploration in the data. Exploratory midtraining~\cite{2611.33130} puts exploration into the demonstration data, collected by people before training. Its online step learns only from successful episodes, mixed with part of the original data.

\section{Problem statement}\label{sec:3}

A VLA policy pretrained on demonstrations is fine-tuned with online RL on a pool of tasks. The reward is sparse: 1 at success, 0 otherwise. For some tasks the pretrained policy succeeds often. For others its success rate is zero or close to zero. We call these unsolved tasks.

Exploration has to produce the first success on an unsolved task. Our question is how to explore when no solution lies close to the current behaviour.

A second requirement comes from lifelong operation. The robot keeps its old tasks while it learns new ones. Exploration for new solutions must leave working skills intact.

\section{Method}\label{sec:4}

\subsection{Base}\label{sec:4.1}

The policy is GR00T N1.7, a later version of GR00T N1~\cite{2503.14734}. Its action head, the part that outputs the actions, is conditioned on the output of the vision-language model and produces actions by flow matching. We call one pass of the policy that outputs one action chunk a call. We follow the public RL recipe for this model~\cite{rlinf}. It uses proximal policy optimisation~\cite{1707.06347}, a policy-gradient method, with noise injected into the sampling of the action head. Our runs of the recipe update two parts: the action head (1.62 billion parameters) and a value head. The vision-language model receives no updates. The value head is a small network that estimates the expected return.

\subsection{Three signals}\label{sec:4.2}

All three signals are computed from the policy's own activations and outputs. Each signal is standardised with exponential moving averages of its mean and standard deviation. It is then compared with a threshold of zero. A standardised signal is zero where it equals its running mean. Novelty and entropy count as high at or above the threshold. Competence counts as low below it.

Novelty is the prediction error of random network distillation (Section~\ref{sec:2}). Its predictor and its fixed random network both read the pooled output of the vision-language model, joined with the state of the robot. The error $e_t$ is standardised: $z_t$ = ($e_t$ - $a_t$) / $s_t$. The running mean $a_t$ and standard deviation $s_t$ are exponential moving averages. Both networks receive inputs limited to five standard deviations. When the statistics are updated, an error above its running mean plus five standard deviations counts as that bound.

Competence is the value head's estimate for the current state.

Action-chunk entropy. Action entropy is estimated from several action chunks sampled for the same observation~\cite{2510.09459}. In the measurements of this paper we sample eight chunks. Our measure is the mean, over the action dimensions, of the logarithm of the standard deviation of the first action across the eight samples. The eight samples share one pass of the vision-language model. In a familiar state with a competent policy, a wide spread means that several solutions exist. In a new state it means that the policy does not know what to do.

Each comparison gives one bit, and the three bits give eight combinations, which we call cells.

\subsection{Short-term exploration: the noise scale}\label{sec:4.3}

Table 1 gives a multiplier on the noise scale for each cell. The multiplier is largest when the state is new, competence is low and the sampled actions spread widely. It is smallest when the state is familiar, the policy is competent and its actions agree. A familiar state with low competence gets less noise than a new state with low competence. Where the policy knows the state and still fails, more random variation is unlikely to help.

\begin{table}[t]
\renewcommand{\thetable}{\arabic{table}}\setcounter{table}{0}
\centering
\footnotesize
\setlength{\tabcolsep}{4pt}
\caption{Multiplier on the noise scale, by cell.}
\label{tab:mult}
\begin{tabular}{lcccc}
\hline
 & \shortstack{competent\\and familiar} & \shortstack{competent\\and new} & \shortstack{low\\competence\\and familiar} & \shortstack{low\\competence\\and new}\\
\hline
Entropy low & 0.4 & 0.7 & 0.8 & 1.0\\
Entropy high & 0.7 & 1.2 & 1.0 & 1.4\\
\hline
\end{tabular}
\end{table}

The table gives eight anchor values, and the multiplier moves smoothly between them. Each standardised signal u becomes a strength q = clip(0.5 + u / 2, 0, 1). Here clip(x, lo, hi) limits x to the range from lo to hi. It is one half at the threshold and reaches 0 and 1 at one standard deviation below and above it. Each cell gets a weight, the product of three factors, one per signal. For each signal, the factor is q if the cell marks that signal high. Otherwise it is 1 - q. The multiplier $g_t$ is the average of the eight values under these weights. It therefore lies between 0.4 and 1.4. The noise scale of the action head at call t is $\sigma_t = \sigma_0\, g_t$. Here $\sigma_0$ is the noise scale of the recipe.

\subsection{Long-term exploration: the novelty reward}\label{sec:4.4}

The same novelty value enters the reward:

\begin{equation}
r'_t = r_t + \beta\, n_t, \qquad n_t = \max(z_t, 0).
\end{equation}

Here $r_t$ is the task reward and $\beta$ is the weight of the novelty term, set as described below. The sum $r'_t$ is the reward that RL training uses. The value head learns the return of $r'$.

Since $z_t$ uses running statistics, $n_t$ measures novelty against the recent past. It shrinks when a state becomes familiar. A state is paid as long as it is more novel than the recent average. We keep paying the novelty reward after a task is solved, for two reasons. A faster solution may exist even when the success rate is at its maximum. The environment of a real robot also keeps changing.

The task reward has to outweigh the novelty term. Otherwise a robot could drop an object or leave the workspace because that is new. The novelty return of an episode is the sum of $\beta$ * $n_t$ over its calls. The weight $\beta$ is set so that the mean novelty return equals one tenth of the reward of one success. A policy that gives up successes for novelty loses return.

The weight $\beta$ is set once, at the start of training. The running statistics are initialised from the first batch of rollouts. We tested whether the novelty return then stays near its target. The test uses 900 stored episodes, ten for each task of LIBERO-90 (Section~\ref{sec:6}). They come from the released GR00T N1.7 checkpoint fine-tuned on the Long suite of LIBERO (Section~\ref{sec:6}), which we call the Long checkpoint. We replayed their calls in 205 rounds of 128 calls each, the size of one rollout batch in training. Each round updates the running statistics and takes four gradient steps of the predictor on the calls of the round. In the first round the novelty return is at its target by construction. Averaged over the next four rounds, the projected novelty return of an episode is 0.04 to 0.05 of the reward of one success. Averaged over the later rounds it is 0.08 to 0.09, close to its target of one tenth. The projection is $\beta$ times the mean novelty of a round times 32, the number of calls in an episode that runs to the horizon. With statistics started at zero mean and unit variance instead, the projection averages 2.3 to 2.9 times the reward of one success from the fifth round on.

\section{Hypotheses and evaluation}\label{sec:5}

Setting. The main evaluation uses two benchmark suites in simulation. LocoManiLab~\cite{2610.30876} has 234 household tasks for the Unitree G1 humanoid, which walks and manipulates. In RobotAgentBench~\cite{2610.34649}, a mobile robot with two arms must also find its household tasks and recover from failures. For the full paper we run both in Isaac Sim and port LocoManiLab to it. LIBERO (Section~\ref{sec:6}) serves only for fast preliminary tests; the full paper adds no LIBERO experiments. Tasks with demonstrations enter supervised fine-tuning before RL; the others are first met in RL. In each suite, all training tasks are trained together in one RL run with randomised initial placements. Success is recorded per task from the training rollouts.

Conditions. (a) Fixed noise. (b) The noise controller of~\cite{2609.28838}, with novelty and competence. (c) The noise scale of Section~\ref{sec:4.3}, with three signals. (d) Condition (c) with the novelty reward. This is the full method, Venture. (e) Condition (a) with the novelty reward. Every condition has three training seeds. Condition (e) separates the effect of the reward from the effect of the noise scale. One ablation runs condition (e) with $\beta$ at half and at double its value. Another runs condition (d) with competence read from an ensemble of value heads, as in~\cite{2609.28838}.

H1 (main). On unsolved tasks, RL with the novelty reward reaches a higher success rate than RL with noise alone. The comparisons are (d) against (c), and (e) against (a).

H2 (retention). On tasks the policy already solves, RL with the novelty reward keeps success at least as high as RL with noise alone. The comparisons are the same two.

H3. Condition (c) reaches a higher success rate on unsolved tasks than condition (b), at equal success on solved tasks.

If H1 or H2 fails. Suppose the visited states become more novel and H1 still fails. Then reaching new states falls short of solving new tasks. The novelty signal would then have to measure learning progress, the rate at which the prediction error falls, instead of the error itself. If H1 fails and the visited states do not become more novel, the reward was too weak to move the policy to new states. Then $\beta$ has to rise. If H1 holds and H2 fails, long-term exploration and retention conflict. Competence then has to switch the novelty term off where the policy already succeeds.

\section{Preliminary measurements}\label{sec:6}

These are fast preliminary tests, on LIBERO unless stated; the main evaluation is on the two suites of Section~\ref{sec:5}.

Success before RL. LIBERO~\cite{2306.03310} is a benchmark of simulated manipulation with one robot arm. Four of its suites have ten tasks each. They vary the layout (Spatial), the objects (Object), the goal within one scene (Goal) and the length of the task (Long). LIBERO-90 adds 90 further tasks. The model's authors released four GR00T N1.7 checkpoints, each fine-tuned on one suite. We ran each checkpoint on all four suites. Every task had ten random placements of its objects, and noise was injected in sampling. Each checkpoint succeeds on its own suite and fails on the others (Table 2). Run three times on these placements, the Long checkpoint solved 74, 75 and 67 of 100; the Goal checkpoint, run twice, solved 82 and 81. Table 2 gives 74 and 81. In the original MuJoCo release of LocoManiLab, we ran the GR00T N1.7 checkpoint that its authors fine-tuned on their atomic training tasks, with their evaluation settings. On the 25 unseen atomic tasks of the kitchen and the living room, it succeeds in 33 of 250 episodes, and 17 of these tasks have no success in ten episodes. The authors report 11.3\% over all 46 unseen atomic tasks~\cite{2610.30876}.

\begin{table}[t]
\renewcommand{\thetable}{\arabic{table}}\setcounter{table}{1}
\centering
\caption{Success out of 100 episodes. Each row is one checkpoint: its success on its own suite, then on each of the other three suites of ten tasks.}
\label{tab:start}
\begin{tabular}{lcc}
\hline
Checkpoint & Own suite & Each other suite\\
\hline
Spatial & 95 & 0\\
Object & 85 & 0\\
Goal & 81 & 0\\
Long & 74 & 0 to 2\\
\hline
\end{tabular}
\end{table}

LIBERO-90 is the task pool of the RL runs below. On it, the Long checkpoint succeeds in 143 of 900 episodes, a rate of 0.159. The 95\% Wilson score interval is 0.136 to 0.184. Of the 90 tasks, 65 have no success in ten episodes, 9 succeed one to four times, and 16 succeed in five or more.

Novelty on solved and unsolved tasks. For this test, a task is solved if it succeeds in five or more of ten episodes. It is unsolved if it has zero successes in the ten episodes. For the Long checkpoint, 26 tasks are solved: the ten Long tasks and 16 of LIBERO-90. The 85 unsolved tasks are 65 of LIBERO-90 and the 20 tasks of Goal and Object, on which the Long checkpoint has no success. Left out are the nine LIBERO-90 tasks with one to four successes and the Spatial tasks, whose stored episodes lack the predictor's input.

A novelty predictor was fitted on rollouts of the solved tasks. We then ask how often a rollout of an unsolved task receives a higher novelty score than a held-out rollout of a solved task. Over three fitting seeds this share is 0.82, 0.84 and 0.83, where 0.5 is chance. For the first fitting seed, the 95\% interval is 0.75 to 0.89, from a bootstrap over tasks. With the labels solved and unsolved permuted across tasks 200 times, the 95th percentile of the share is 0.61. Most unsolved tasks come from suites new to the checkpoint, so the score could reflect the suite alone; within LIBERO-90 alone, the share is 0.81 to 0.89.

With the input of the action head at its last sampling step as the predictor's input, the score fails to separate solved from unsolved tasks within LIBERO-90 (0.44 to 0.46). This is why novelty is read from the output of the vision-language model. On recorded demonstrations of LocoManiLab, not rollouts of the policy, a predictor fitted on the training tasks of that checkpoint scores an episode of one of the 46 unseen tasks above a held-out episode of a training task with a share of 0.91 to 0.96 across nine fits, and 0.61 with permuted labels.

A second model. We repeated the novelty test with $\pi_{0.5}$~\cite{2504.16054}, a second VLA policy. The checkpoint is public and was fine-tuned on four LIBERO suites~\cite{rlinf}. The feature is the mean output of its vision-language model, with the robot state appended. Two feature dimensions held non-finite values and were dropped by a rule fixed before the results. The checkpoint solves 165 of 900 episodes of LIBERO-90, and its 57 unsolved tasks are all LIBERO-90 tasks. Within LIBERO-90, the share is 0.95 to 0.97 over three fitting seeds. Over all suites the share is 0.97 to 0.98, and with the labels permuted, the 95th percentile of the share is 0.58 to 0.59. Across suites the solved and unsolved rollouts also differ in sampler settings, which is why we give the LIBERO-90 share first. The rollouts use one set of placements.

Entropy inside the cells. We logged the three signals on 100 episodes of the Long checkpoint on its own suite. The first 512 calls only set the running statistics, and the 1597 calls after them were scored. For every call we took the success of its episode. In each of the four combinations of novelty and competence, calls with high entropy belong to less successful episodes than calls with low entropy (Table 3). On Long, each 95\% interval, computed over episodes, excludes zero. On 300 episodes of 30 LIBERO-90 tasks, where success is 0.15, the differences are smaller. The interval for low competence in a new state includes zero. In the competent and familiar cell this goes against the reading of Section~\ref{sec:4.2}, which took a wide spread there as a sign of several solutions. Table 1 gives that cell more noise at high entropy, and training has to show whether this choice holds.

\begin{table}[t]
\renewcommand{\thetable}{\arabic{table}}\setcounter{table}{2}
\centering
\footnotesize
\setlength{\tabcolsep}{4pt}
\caption{Episode success of calls with high entropy minus that of calls with low entropy, in the four combinations of novelty and competence.}
\label{tab:3}
\begin{tabular}{lcccc}
\hline
 & \shortstack{competent\\and familiar} & \shortstack{competent\\and new} & \shortstack{low\\competence\\and familiar} & \shortstack{low\\competence\\and new}\\
\hline
Long & $-$0.217 & $-$0.205 & $-$0.223 & $-$0.200\\
LIBERO-90 & $-$0.045 & $-$0.041 & $-$0.060 & $-$0.006\\
\hline
\end{tabular}
\end{table}

Competence. Competence from the single value head says little about episode success. In the two familiar cells with low entropy, episode success on Long is 0.66 with high competence and 0.66 with low competence. On LIBERO-90 it is 0.12 and 0.11.

The smooth multiplier. For each signal, 69 to 71 percent of the Long calls lie within one standard deviation of the threshold. For these calls the multiplier of Section~\ref{sec:4.3} blends values of Table 1. Averaged over all scored calls, it differs from the Table 1 value of the call's cell by 0.12. The measurements of entropy, competence and the smooth multiplier come from one checkpoint and one run per suite.

RL with and without the novelty reward. Two runs train on LIBERO-90 from the Long checkpoint. The run with the novelty reward is condition (e) of Section~\ref{sec:5}, and the run without it is condition (a). Conditions (b), (c) and (d), which adapt the noise scale, have no runs in this paper. The ten Long tasks are held out from training. An epoch is one batch of rollouts and the updates on it, and a run has 460 epochs. Success per task was counted up to epoch 365 in the run with the reward and up to epoch 355 in the run without it. With the reward, the 65 tasks with zero success at the start succeed in 3 of 1156 training episodes. Without the reward, the same 65 tasks succeed in 5 of 1106. The reward shows no gain on these tasks.

We evaluate both runs on the ten Long tasks at four epochs, with 100 episodes each time (Table 4). On the same placements, the starting checkpoint solves 74. We compare each count with that of the starting checkpoint by the exact McNemar test for paired outcomes. With the reward, no count differs from the starting checkpoint at the 0.05 level. Without the reward, the counts at epochs 184 and 368 are lower. Between the two runs at epoch 460, the test gives p = 0.12. Two further runs train condition (e) for 92 epochs, one with half and one with double $\beta$. At epoch 92, the run with half $\beta$ solves 67 and the run with double $\beta$ 73, on the same placements.

We repeated the evaluation at epoch 92 with three evaluation seeds. Each seed draws a new set of 100 placements, and all runs share them. The first seed is the set of Table 4. On these 300 episodes, the starting checkpoint solves 231. Against the starting checkpoint, the run with the reward solves 215 (p = 0.068) and the run without it 199 (p = 0.0005). The runs with half and double $\beta$ solve 209 and 218 (p = 0.020 and 0.15). With one training seed per run, these counts are observations rather than tests of the hypotheses.

\begin{table}[t]
\renewcommand{\thetable}{\arabic{table}}\setcounter{table}{3}
\centering
\footnotesize
\setlength{\tabcolsep}{4pt}
\caption{Success out of 100 episodes on the ten Long tasks, where the starting checkpoint solves 74. In brackets, the p value against the starting checkpoint. Columns: epochs 92, 184, 368, 460.}
\label{tab:4}
\begin{tabular}{lcccc}
\hline
Epoch & 92 & 184 & 368 & 460\\
\hline
With the reward & 73 (1.00) & 73 (1.00) & 68 (0.31) & 75 (1.00)\\
Without the reward & 63 (0.061) & 60 (0.016) & 59 (0.014) & 66 (0.15)\\
\hline
\end{tabular}
\end{table}

\section{Limits}\label{sec:7}

All measurements are in simulation.

\section*{Use of language models}
Large language models were used to draft and edit the text of this paper from the author's written notes and answers, and to write and review code. The author read and checked all content and is responsible for it.

\bibliographystyle{plainnat}
\bibliography{references}

\end{document}